%% file: main_iclr.tex
\documentclass{article}
\usepackage{iclr2024_conference,times}

\input{math_commands.tex}

\usepackage{amsmath}
\usepackage{amssymb}
\usepackage{natbib}
\usepackage{float}
\usepackage{tabu}
\usepackage{algorithm}
\RequirePackage{algorithmic}
\usepackage{fontawesome5}
\usepackage{hyperref}
\usepackage{url}
\usepackage[utf8]{inputenc} 
\usepackage{url}           
\usepackage{booktabs}       
\usepackage{amsfonts}       
\usepackage{nicefrac}       
\usepackage{microtype}      
\usepackage{xcolor}         
\usepackage{tcolorbox}

\usepackage{graphicx}
\usepackage{longtable}
\usepackage{caption}
\usepackage{mdframed}
\usepackage{subcaption}
\usepackage{multirow}
\usepackage{placeins}
\usepackage{multicol}
\usepackage{makecell}
\usepackage{soul}
\usepackage[normalem]{ulem} 
\usepackage{wrapfig}
\usepackage[percent]{overpic}
\usepackage{lipsum}
\usepackage{csquotes}
\usepackage[OT2,T1]{fontenc}
\usepackage[english]{babel}
\usepackage{devanagari}
\usepackage{tablefootnote}
\usepackage{pdfpages}
\usepackage{colortbl}
\usepackage{arydshln}
\usepackage{array}
\usepackage{enumitem}

\definecolor{myblue}{RGB}{0, 0, 255}
\definecolor{mygray}{gray}{0.9}

\newcommand{\myparagraph}[1]{{\bf #1}}

\newcommand{\ie}{\textit{i}.\textit{e}.}
\newcommand{\eg}{\textit{e}.\textit{g}.}
\newcommand{\etc}{\textit{etc}}

\usepackage{pifont}

\newcommand{\cmark}{\ding{52}}
\newcommand{\fmark}{\ding{56}}
\def\halfcheckmark{\textcolor{black}{\ding{52}}{\small\textcolor{black}{\kern-0.7em\ding{55}}}}

\newcommand{\authorskip}{\hspace{4.8mm}}

\definecolor{myGreen}{rgb}{0, .6, .0}
\definecolor{goodblue}{HTML}{0071bc}
\hypersetup{colorlinks=true,breaklinks,linkcolor=red,citecolor=goodblue}

\title{Diffusion Feedback Helps CLIP See Better}

\usepackage{xspace}
\def\ours{\textbf{DIVA}\xspace}

\author{
Wenxuan Wang\textsuperscript{1,2,3}\thanks{Equal contribution. $\dag$ Correspondence to \textit{wangxinlong@baai.ac.cn}.} \hspace{-1.0em}
\authorskip Quan Sun\textsuperscript{3}$^*$ \hspace{-1.5em}
\authorskip Fan Zhang\textsuperscript{3} \hspace{-1.5em}
\authorskip Yepeng Tang\textsuperscript{4} \hspace{-1.5em}
\authorskip Jing Liu\textsuperscript{1,2} \hspace{-1.5em}
\authorskip Xinlong Wang\textsuperscript{3}$^\dag$ \\[2mm]
\textsuperscript{1} Institute of Automation, Chinese Academy of Sciences \hspace{5.5mm} \\
\textsuperscript{2} School of Artificial Intelligence, University of Chinese Academy of Sciences \hspace{5.5mm} \\
\textsuperscript{3} Beijing Academy of Artificial Intelligence \hspace{5.5mm} \\
\textsuperscript{4} Institute of Information Science, Beijing Jiaotong University \\[2.5mm]
{
\centerline{
Project Page: \textit{\url{https://rubics-xuan.github.io/DIVA/}}
}
}
}

\iclrfinalcopy
\begin{document}

\maketitle

\begin{abstract}

Contrastive Language-Image Pre-training (CLIP), which excels at abstracting open-world representations across domains and modalities, has become a foundation for a variety of vision and multimodal tasks.
However, recent studies reveal that CLIP has severe visual shortcomings, such as which can hardly distinguish orientation, quantity, color, structure, \etc.
These visual shortcomings also limit the perception capabilities of multimodal large language models (MLLMs) built on CLIP.
The main reason could be that the image-text pairs used to train CLIP are inherently biased, due to the lack of the distinctiveness of the text and the diversity of images.
In this work, we present a simple post-training approach for CLIP models, which largely overcomes its visual shortcomings via a self-supervised diffusion process.
We introduce \ours, which uses the \textbf{{D}}\textbf{{I}}ffusion model as a \textbf{{V}}isual \textbf{{A}}ssistant for CLIP.
Specifically, \ours leverages generative feedback from text-to-image diffusion models to optimize CLIP representations, with only images (without corresponding text). 
We demonstrate that \ours improves CLIP's performance on the challenging MMVP-VLM benchmark which assesses fine-grained visual abilities to a large extent (\eg, 3-7\% ↑), and enhances the performance of MLLMs and vision models on multimodal understanding and segmentation tasks. 
Extensive evaluation on 29 image classification and retrieval benchmarks confirms that our framework preserves CLIP's strong zero-shot capabilities. 
The code will be available at \textit{\url{https://github.com/baaivision/DIVA}}.

\end{abstract}

\section{Introduction}

Contrastive language-image pre-training (CLIP) \citep{radford2021learning} has been widely applied to various multimodal understanding and generation tasks, including open-domain image classification \citep{sun2024alpha,zhang2022tip,zhu2023not}, text-to-image retrieval \citep{luo2023lexlip,baldrati2022effective,sain2023clip}, 
visual grounding \citep{wang2022cris,yu2023zero,wang2024unveiling,wang2024cm},
and text-to-image generation \citep{frans2022clipdraw,bar2022text2live,rombach2022high,crowson2022vqgan,ramesh2022hierarchical,vinker2022clipasso}. 
This widespread application is due to CLIP's excellent visual representation ability, learned from large-scale data. Thus, enhancing CLIP's representation and capabilities is crucial for advancing downstream tasks.

Since the introduction of CLIP \citep{radford2021learning}, numerous subsequent studies on CLIP models have emerged in recent years. 
These studies have utilized training techniques pre-train \citep{sun2023eva,sun2024eva18b,fang2023data, xu2023demystifying, zhai2023sigmoid,shi2024umg} and fine-tune \citep{wei2023improving,zhang2024long} CLIP models, achieving improved performance and unlocking new abilities. However, these approaches still suffer from unavoidable limitations, as they heavily rely on image-text data pairs and cannot work on image-only data.

\begin{figure}[t]
  \centering
  \includegraphics[width=0.92\textwidth]{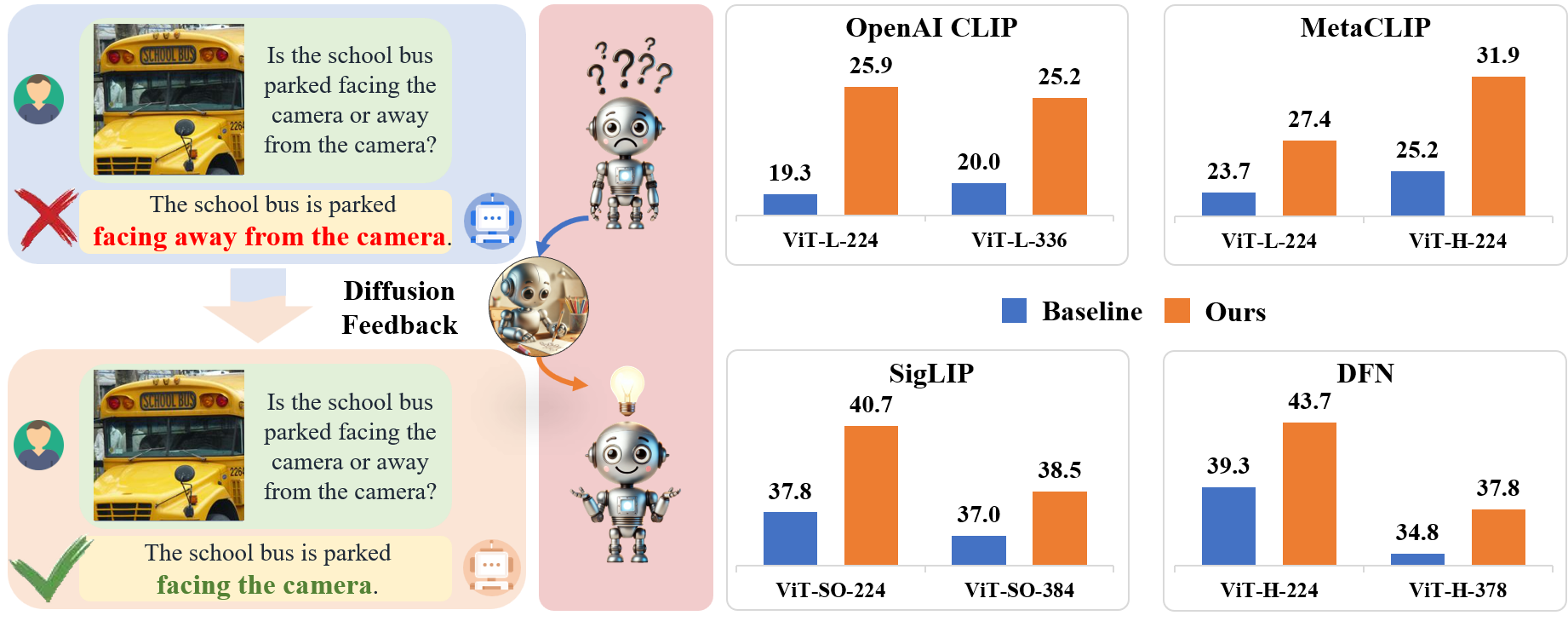}
  \vspace{-10pt}
  \caption{\textbf{Left:} The existing CLIP models mostly suffer from the inability to distinguish visual details. After enhancing the visual capabilities with our \ours, the sensitivity of CLIP to visual details has greatly improved. \textbf{Right:} Our proposed \ours consistently boosts the performance of various CLIP models~\citep{radford2021learning, fang2023data, xu2023demystifying, zhai2023sigmoid} on MMVP-VLM benchmark that evaluates the visual capabilities of vision-language models.}
  \label{fig:introduction}
  \vspace{-15pt}
\end{figure}

As noted by recent works \citep{kim2023region, zeng2021multi, zhang2024long, tong2024eyes, tong2024cambrian}, despite its excellent zero-shot performance, CLIP suffers from certain perceptual understanding limitations due to the contrastive learning paradigm and the noisy image-text pairs used in training. These limitations include an inability to accurately comprehend long texts and to perceive fine-grained differences in similar images. While some studies have attempted to address the text comprehension issue \citep{zhang2024long}, research on improving CLIP's fine-grained visual perception remains underexplored. The ability to perceive visual detail is crucial for foundation models, and the lack of this capability in CLIP directly affects the performance of vision and multimodal models that use CLIP as a vision encoder \citep{tong2024eyes, tong2024cambrian}.

Thus, in this work, we focus on addressing CLIP's inability to distinguish fine-grained visual details via self-supervised learning (SSL) paradigm. 
Based on text-to-image diffusion models generating  realistic images with much details, we explore leveraging generative feedback from diffusion models to optimize CLIP representations. 
By conditioning diffusion models with CLIP's densely recapped visual features and applying reconstruction loss to CLIP optimization, 
we leverage the \textbf{{D}}\textbf{{I}}ffusion model as a \textbf{{V}}isual \textbf{{A}}ssistant for CLIP, thus the name of our approach, \ours.
Our results highlight that \ours greatly enhances CLIP's performance on MMVP-VLM benchmark measuring visual abilities of V-L models, and improves multimodal large language models (MLLMs) and vision models on multimodal and vision understanding tasks.
Besides, \ours maintains CLIP's excellent zero-shot performance on 29 image classification and retrieval benchmarks.

Our main contributions can be summarized as follows:
\setlist{nolistsep}
\begin{itemize}[noitemsep,leftmargin=*]
    \item 
    Concentrating on overcoming CLIP's visual shortcomings in perceiving fine-grained details, we present the first work to exploit the potential of leveraging generative feedback from text-to-image diffusion models to optimize CLIP model's discriminative representations.
    \item 
    We propose a simple self-supervised framework \ours for CLIP's representation optimization. Coupled with our visual dense recap scheme, \ours conditions diffusion models with dense visual features from CLIP and incorporates image reconstruction loss for optimization.
    \item 
    Our \ours greatly boosts CLIP's visual perception capability and improves its performance on MMVP-VLM benchmark, further enhancing MLLMs and vision models on multimodal and visual understanding tasks. Meanwhile, our results on 29 image classification and retrieval benchmarks show that \ours maintains CLIP's original excellent zero-shot performance.
\end{itemize}

\vspace{-5pt}
\section{Related Work}
\vspace{-5pt}
\myparagraph{CLIP Models \& MLLMs.}
The introduction of CLIP \citep{radford2021learning} has significantly advanced multimodal learning. Since its debut, a series of CLIP models have emerged \citep{sun2023eva,fang2023data, xu2023demystifying, zhai2023sigmoid}, enhancing performance and unlocking new capabilities through improved pre-training techniques and model architectures. 
On this basis, CLIP has been widely adopted as a foundation model, serving as a backbone for various applications such as image segmentation \citep{li2022language,xu2022groupvit,shan2024open,xu2023side,liang2023open,zhou2023zegclip}, object detection \citep{gu2021open,li2022grounded,subramanian2022reclip} and video understanding \citep{bose2023movieclip,lin2022frozen,castro2022fitclip,xu2021videoclip,rasheed2023fine,tang2024learnable}. Its ability to align language and vision has led to superior results on these tasks compared to traditional methods.
Moreover, CLIP has driven the development of MLLMs \citep{liu2024visual,liu2024improved,sun2024emu,sun2024emu2}. Combining strong visual understanding with advanced large language models facilitates more sophisticated interactions between vision and language. Recent works have highlighted inherent visual flaws in the CLIP models and MLLMs using CLIP as the visual encoder \citep{tong2024eyes, tong2024cambrian}. To address this, some research has incorporated multiple vision encoders to achieve more precise and comprehensive visual perception \citep{kar2024brave, jiang2023clip, tong2024eyes}. However, this approach increases computational costs and memory usage. There has been no research directly enhancing CLIP's visual perception capabilities to better serve MLLMs. Thus, the main focus of our work is to fundamentally overcome CLIP's visual perception shortcomings, directly benefiting both vision models and multimodal MLLMs that use CLIP as a backbone.

\myparagraph{Diffusion Models for Representation Learning.}
Diffusion models~\citep{ho2020denoising,song2020score} have made remarkable progress in various generative tasks, such as image generation~\citep{stablediffusion, imagen, dalle3, cogview3}, video generation~\citep{makeavideo, svd, show1, imagenvideo}, editing~\citep{sdedit, instructimagen, dragondiffusion}, etc. 
Apart from research above, there are also many works focus on employing diffusion models for representation learning. 
Some of works leverage the intermediate activation of pre-trained diffusion models for different downstream tasks, including classification~\citep{ddae}, semantic segmentation~\citep{ddpmseg}, panoptic segmentation~\citep{ODISE}, depth estimation~\citep{VDM}, etc. 
Other works~\citep{soda, MDM} train their own diffusion models coupled with meticulously devised modules to further boosting the representation capabilities. 
Besides, Diffusion-TTA~\citep{prabhudesai2023diffusion} aims to adapt pre-trained vision encoders to samples in testing set using feedback from a diffusion model. 
Additionally, some methods~\citep{guo2024everything, trabucco2023effective, tian2024stablerep, azizi2023synthetic} utilize diffusion models to generate synthetic data, which is then adopted to enhance the representation capabilities of corresponding models. 
In contrast, in our work, we mainly focus on exploring the potential of enhancing the original discriminative representations of CLIP models by directly utilizing generative feedback from the diffusion models. 
Additionally, we aim to leverage the diffusion models to break free from the constraints of paired image-text data and construct a self-supervised framework to improve CLIP's visual perception capabilities.

\newcommand{\bzero}{\mathbf{0}}
\newcommand{\bone}{\mathbf{1}}
\newcommand{\cond}{\textbf{c}}

\vspace{-5pt}
\section{Enhancing CLIP's Representations via Diffusion Feedback}
\vspace{-5pt}

In this section, we present our \ours, an effective framework for boosting CLIP's visual perception capabilities with a pre-trained conditional diffusion model. 
We first discuss CLIP's visual deficiencies in perceiving details and generative diffusion models as preliminaries in Sec. \ref{subsec:clip_prelim} and Sec. \ref{subsec:diffusion_prelim} respectively.
Then the overall architecture of \ours is illustrated in Sec. \ref{subsec:overall}, followed by our well-designed visual dense recap strategy for better unleashing the power of \ours in Sec. \ref{subsec:diff_condition}.

\vspace{-5pt}
\subsection{Preliminaries}
\subsubsection{CLIP's Visual Deficiencies}
\label{subsec:clip_prelim}

Thanks to the excellent representations learned through pre-training on massive data, CLIP \citep{radford2021learning} exhibits outstanding generalization capabilities and is widely applied in the V-L understanding domain. 
However, CLIP is not without its flaws. 
As highlighted in the study \citep{tong2024eyes}, CLIP struggles to distinguish detailed differences between two images that are obviously distinct to human observers. 
This deficiency mainly stems from two aspects:
1) \textbf{Training Paradigm}: The original contrastive learning strategy of CLIP aims to minimize the distance between positive pairs and maximize the distance between negative pairs of visual class tokens and textual semantics, 
resulting in visual perception bias that mainly focuses on high-level semantic information while overlooking visual details such as orientation, quantity, color, and structure. 
Consequently, CLIP sometimes encodes visually different images into similar embeddings, making it difficult to differentiate these images' subtle variations.
2) \textbf{Data Format}: The text in the image-text pairs used to train CLIP is limited in length. As pointed out by \citep{zhang2024long}, although the length of the text token is restricted to 77, CLIP's actual effective text length is less than 20. Therefore, the textual data in these image-text pairs inherently lacks descriptions of the visual details in the corresponding positive sample images. This fundamental limitation of the training data also leads to CLIP's inability to adequately perceive visual detail information.

\subsubsection{Generative Diffusion Models}
\label{subsec:diffusion_prelim}

A diffusion model learns to model a probability distribution by reversing a process that progressively adds noise to the image.
Given an image sample $x_0$ drawn from an underlying probability distribution $p(x)$, a forward diffusion process defines a Markov chain to gradually add random Gaussian noise $\mathbf{\epsilon_t} \in \mathcal{N}(\mathbf{0}, \mathbf{I})$ to the original sample $\mathbf{x_0}$:
\begin{equation}
\label{e:diff_xt_xt-1}
\mathbf{x_{t}} = \sqrt{1-\beta_t} \mathbf{x_{t-1}} + \sqrt{\beta_t} \mathbf{\epsilon_t},\quad t =1,\ldots,T
\end{equation}
Here, $T$ denotes the number of diffusion steps, and $\beta_t \in (0, 1)$ is a predefined time-dependent variance schedule. As the $T$ becomes large enough, $\mathbf{x_T}$ is close to $\mathcal{N}(\mathbf{0}, \mathbf{I})$.
The transition equation can be reformulated as follows by leveraging the additive property of Gaussian distribution:
\begin{equation}
\label{e:diff_xt_x0}
\mathbf{x_{t}} = \sqrt{\bar{\alpha}_t} \mathbf{x_0} + \sqrt{1-\bar{\alpha}_t} \mathbf{\epsilon},\quad t =1,\ldots,T
\end{equation}
in which $\alpha_t=1-\beta_t$ and $\bar{\alpha}_t=\prod_{i=1}^{t}\alpha_i$.
On this basis, the image sample $x_0$ can be iteratively generated from a random noise $\mathbf{x_T} \sim \mathcal{N}(\mathbf{0}, \mathbf{I})$ by reversing the forward diffusion process:
\begin{equation}
\label{e:diff_reverse}
    \mathbf{x_{t-1}} = \frac{1}{\sqrt{\alpha_t}}(\mathbf{x_t} - \frac{1-\alpha_t}{\sqrt{1-\bar{\alpha}_t}}\mathbf{\epsilon_\phi}(\mathbf{x_t}, t)) + \sigma_t \mathbf{\epsilon},\quad t =T,\ldots,1
\end{equation}
$\mathbf{\epsilon_\phi}$ is a denoising neural network trained to predict $\mathbf{\epsilon}$ in the forward diffusion process and $\sigma_t$ is the posterior noise variance.
A commonly used training objective for a diffusion model $\mathbf{\epsilon_\phi}$ is:
\begin{equation}
\label{e:diff_loss_uncond}
L(\phi)=\mathbb{E}_{t,\mathbf{x_0},\epsilon}\left[\|\mathbf{\epsilon} - \mathbf{\epsilon_\phi}(\sqrt{\bar{\alpha}_t}\mathbf{x_0}+\sqrt{1-\bar{\alpha}_t}\mathbf{\epsilon}, t) \|^2 \right]
\end{equation}
Besides, the diffusion models can be easily extended to conditional generation by incorporating a condition $\mathbf{c}$ into $\mathbf{\epsilon_\phi}$, in which $\mathbf{c}$ can be a class label, a text prompt or a reference image. Thus, the training objective should be modified to:
\begin{equation}
\label{e:diff_loss_cond}
    L(\phi)=\mathbb{E}_{t,\mathbf{x_0},\epsilon,\mathbf{c}}\left[\|\mathbf{\epsilon} - \mathbf{\epsilon_\phi}(\sqrt{\bar{\alpha}_t}\mathbf{x_0}+\sqrt{1-\bar{\alpha}_t}\mathbf{\epsilon}, t, \mathbf{c}) \|^2 \right]
\end{equation}

\subsection{Overall Structure of \ours}
\label{subsec:overall}

\begin{figure}[t]
  \centering
  \includegraphics[width=0.92\textwidth]{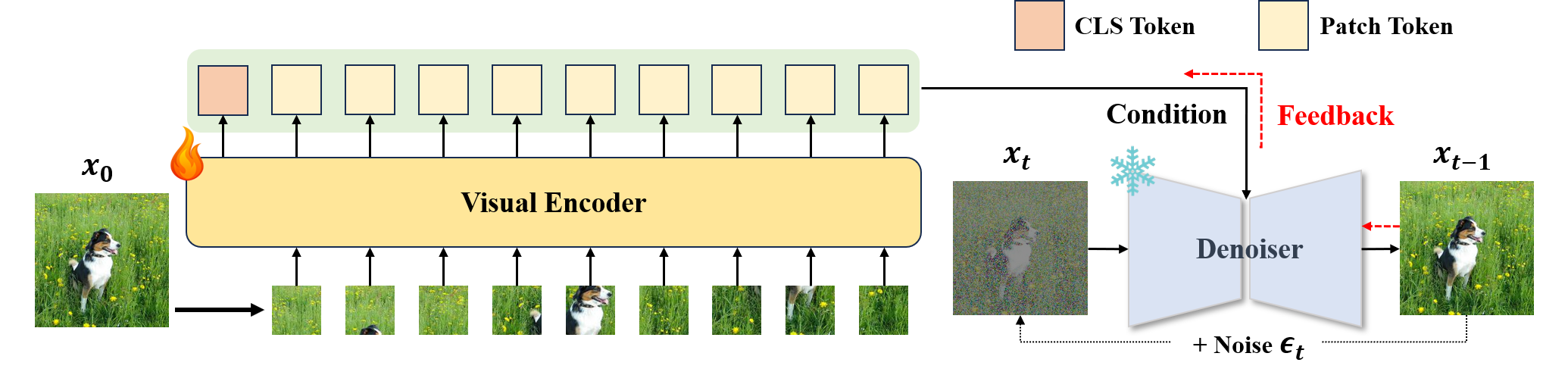}
  \vspace{-12pt}
  \caption{\textbf{Overall architecture of our \ours.} Given an image $x_0$, the CLIP model $\theta$ encodes the visual features as main part of condition $\cond$, then the generative diffusion model $\phi$ predicts the added noise $\epsilon$ taking the noisy image $x_t$ and condition $\cond$ as input. We optimize the CLIP's representation by maximizing the image likelihood with the diffusion loss via generative feedback.}
  \label{fig:methodology}
  \vspace{-15pt}
\end{figure}

As illustrated in Fig. \ref{fig:methodology}, \ours mainly consists of two parts: the CLIP model to be enhanced in terms of visual perception capabilities, and the pre-trained text-to-image diffusion model providing generative feedback. 
Taking an original image as input, the CLIP model encodes the corresponding visual features, which will be combined with the empty text's embeddings (\ie, [BOS] \& [EOS]) from the diffusion model's text encoder for diffusion's condition. 
Given the image with added noise, the diffusion model attempts to predict the noise added from the previous step to the current step with the aforementioned condition. 
This process needs to be repeated N times because, for each image, we will randomly select N states (\ie, two) from the total steps (\eg, 0$\sim$1000 steps) of the diffusion model for optimization.
The corresponding loss function can be represented as Equation \ref{e:diff_loss_cond}. 
Keeping all parts' weights except the CLIP visual encoder frozen, the training objective is simply to minimize the reconstruction loss (\ie, generative guidance). 
In this manner, by constraining the diffusion model to more accurately predict the added noise, the CLIP’s original semantic-rich discriminative representations will be gradually optimized into representations with more visual details through diffusion feedback. On this basis, results in Sec. \ref{subsec:generalization} demonstrate that our \ours does not greatly damage the zero-shot performance of the original CLIP models.
The pseudo code of the specific enhancement process can be found at Algorithm \ref{alg:diff-tta} in \textcolor{blue}{Appendix}.

\subsection{Diffusion's Condition Design}
\label{subsec:diff_condition}
\vspace{-5pt}

\myparagraph{Visual Dense Recap Scheme.}
In our \ours, the diffusion model's condition design is pivotal, as it sets the upper limit for enhancing CLIP's visual capabilities. We introduce a simple yet effective strategy called visual dense recap. Unlike detailed re-captioning of an image's caption in natural language, our approach performs re-captioning at the level of visual richness by incorporating features from local patch tokens along with the class token into the condition. When only the class token is present, CLIP's visual features primarily contain strong semantic information, which is insufficient for reconstructing the original image. Consequently, the reconstruction task becomes challenging due to the lack of adequate information, and CLIP cannot learn significantly enhanced representations. By incorporating local patch features, the auxiliary function of the condition is significantly enhanced, allowing the generative feedback to effectively improve CLIP's visual perception capabilities. We conduct ablation studies in Sec. \ref{sec:ablation} to demonstrate the efficacy of visual dense recap.

\myparagraph{Visual Recap Density.} Although the visual dense recap scheme appears straightforward, the density of the recap is crucial. If the density is too high (\ie, introducing too many local tokens), the richness of the condition information approaches its maximum, greatly reducing the difficulty of reconstruction task. This results in CLIP's representation requiring minimal optimization to easily complete the reconstruction, limiting the upper bound of CLIP's optimized capabilities. Conversely, if the recap density is too low (\ie, retaining only class token or introducing few local tokens), CLIP's optimization process will struggle with the high difficulty of reconstruction, failing to adequately learn the expected detailed visual representations. This intuitive point is confirmed in Sec. \ref{sec:ablation}. 
Specifically, ensuring the visual class token is always present in the condition, we introduce randomly selected local token features with approximately 15\% and 30\% probabilities for OpenAI CLIP \citep{radford2021learning} at 224 and 336 resolutions. 
For the SigLIP ViT-SO-14 \citep{zhai2023sigmoid} at 224 and 384 image sizes, we incorporate local token features obtained through 1D average pooling with local window sizes of 6 and 10, respectively. 
Except that introducing 50\% randomly selected patch tokens into the condition for DFN ViT-H-14/378, for the remaining baselines \citep{fang2023data, xu2023demystifying}, we include all local token features for the condition design.
Apart from DFN ViT-H-14/224 and SigLIP ViT-SO-14/224\&384 \citep{zhai2023sigmoid} only using visual class token, all other models incorporate local features consistent with the training stage conditions during inference, combining them with class token to fully leverage the detailed representations captured by the enhanced CLIP.

\vspace{-5pt}
\section{Experimental Results}
\vspace{-5pt}

To evaluate the effectiveness of our \ours and demonstrate its potential to enhance CLIP representations, comprehensive experiments are conducted on multimodal understanding and visual perception tasks, which will be elaborated in the followings.

\vspace{-5pt}
\subsection{Implementation Details}
\vspace{-5pt}

\ours is trained on 8 NVIDIA-A100 80GB GPUs with a global batch size of 640.
We adopt Stochastic Gradient Descent (SGD) optimizer with a learning rate of 1e-4 and momentum of 0.9 to refine the representations of CLIP models via generative feedback. 
We only optimize the CLIP models with relatively high-quality Conceptual-3M dataset \citep{sharma2018conceptual} for 4600 steps (\ie, nearly 1 epoch) among the training phase, which can already boost CLIP's performance in a training-efficient manner. 
For all experiments, we adjust the parameters of the discriminative CLIP vision encoders and keep the pre-trained diffusion models' parameters frozen through the training process. 

\vspace{-5pt}
\subsection{Fine-grained Visual Perception Evaluation}
\vspace{-5pt}

\definecolor{LightGreen}{HTML}{ccffcc}
\definecolor{Green}{HTML}{99ff99}
\definecolor{DarkGreen}{HTML}{66cc66}

\definecolor{LightRed}{HTML}{ffcccc}
\definecolor{Red}{HTML}{ff9999}
\definecolor{DarkRed}{HTML}{ff6666}

\definecolor{LightBlue}{HTML}{cce0ff}
\definecolor{Blue}{HTML}{99ccff}
\definecolor{DarkBlue}{HTML}{668cff}
\definecolor{LightTeal}{HTML}{B3FFFF}
\definecolor{MediumTeal}{HTML}{66FFFF}
\definecolor{DarkTeal}{HTML}{33CCCC}
\definecolor{LightYellow}{HTML}{FFFFCC}
\definecolor{MediumYellow}{HTML}{FFFF99}
\definecolor{DarkYellow}{HTML}{FFFF66}
\definecolor{lightgray}{gray}{0.9}

\newcommand{\plusvalue}[1]{\hspace{0.3em}\textcolor{darkgreen}{(+#1)}}
\newcommand{\minusvalue}[1]{\hspace{0.3em}\textcolor{red}{(-#1)}}
\definecolor{darkgreen}{rgb}{0.0, 0.5, 0.0}
\definecolor{MidGreen}{HTML}{AAFFAA}

\begin{table}[t]
    \caption{\textbf{Performance of CLIP based models on various visual patterns of MMVP-VLM benchmark.} 
    Our framework greatly overcomes CLIP's original shortcomings in terms of perceiving visual details.
    Symbols for visual patterns as \citep{tong2024eyes} are inherited:
    \textbf{\faCompass}: Orientation and Direction, \textbf{\faSearch}: Presence of Specific Features, \textbf{\faSync}: State and Condition, \textbf{\faSortNumericUp}: Quantity and Count, \textbf{\faMapPin}: Positional and Relational Context, \textbf{\faPalette}: Color and Appearance, \textbf{\faCogs}: Structural and Physical Characteristics, \textbf{\faFont}: Texts, \textbf{\faCamera}: Viewpoint and Perspective.}
    \label{tab:clip_sota}
    \centering
    \small
    \setlength\tabcolsep{1.8pt} 
    \begin{tabular}{l:c:cc:ccccccccc:c}
    \toprule
        \multirow{2}{*}{\shortstack{Method}} & \multirow{2}{*}{\shortstack{Ours}} & \multirow{2}{*}{\shortstack{Image Size}} & \multirow{2}{*}{\shortstack{Params (M)}} & \multirow{2}{*}{\faCompass}  &  \multirow{2}{*}{\faSearch} & \multirow{2}{*}{\faSync} & \multirow{2}{*}{\faSortNumericUp} & \multirow{2}{*}{\faMapPin} & \multirow{2}{*}{\faPalette}  & \multirow{2}{*}{\faCogs}  & \multirow{2}{*}{\faFont} & \multirow{2}{*}{\faCamera}  & \multirow{2}{*}{\shortstack{Average}}
         \\
         \\
        \hline 

        OpenAI ViT-L-14 &  & 224$^2$ & 427.6 & 13.3 & 13.3 & 20.0 & 20.0 & 13.3 & 53.3 & 20.0 & 6.7 & 13.3 & 19.3 \\
        \rowcolor{lightgray} OpenAI ViT-L-14 & \cmark & 224$^2$ & 427.6 & 13.3 & 20.0 & 40.0 & 6.7 & 20.0 & 53.3 & 46.7 & 20.0 & 13.3 & \textbf{25.9}{\textbf{\tiny{\plusvalue{6.6}}}}\\
        OpenAI ViT-L-14 &  & 336$^2$ & 427.9 & 0.0 & 20.0 & 40.0 & 20.0 & 6.7 & 20.0 & 33.3 & 6.7 & 33.3 & 20.0 \\
        \rowcolor{lightgray} OpenAI ViT-L-14 & \cmark & 336$^2$ & 427.9 & 26.7 & 20.0 & 33.3 & 13.3 & 13.3 & 46.7 & 26.7 & 6.7 & 40.0 & \textbf{25.2}{\textbf{\tiny{\plusvalue{5.2}}}} \\
        \hline

        MetaCLIP ViT-L-14 &  & 224$^2$ & 427.6 & 13.3 & 6.7 & 66.7 & 6.7 & 33.3 & 46.7 & 20.0 & 6.7 & 13.3 & 23.7 \\
        \rowcolor{lightgray} MetaCLIP ViT-L-14 & \cmark & 224$^2$ & 427.6 & 6.7 & 6.7 & 60.0 & 0.0 & 26.7 & 66.7 & 20.0 & 20.0 & 40.0 & \textbf{27.4}{\textbf{\tiny{\plusvalue{3.7}}}} \\
        MetaCLIP ViT-H-14 &  & 224$^2$ & 986.1 & 6.7 & 13.3 & 60.0 & 13.3 & 6.7 & 53.3 & 26.7 & 13.3 & 33.3 & 25.2 \\
        \rowcolor{lightgray} MetaCLIP ViT-H-14 & \cmark & 224$^2$ & 986.1 & 13.3 & 20.0 & 53.3 & 33.3 & 13.3 & 66.7 & 33.3 & 13.3 & 40.0 & \textbf{31.9}{\textbf{\tiny{\plusvalue{6.7}}}}\\
        \hline
        
        SigLIP ViT-SO-14 &  & 224$^2$ & 877.4 & 26.7 & 20.0 & 53.3 & 40.0 & 20.0 & 66.7 & 40.0 & 20.0 & 53.3 & 37.8 \\
        \rowcolor{lightgray} SigLIP ViT-SO-14 & \cmark & 224$^2$ & 877.4 & 13.3 & 26.7 & 60.0 & 46.7 & 13.3 & 73.3 & 53.3 & 26.7 & 53.3 & \textbf{40.7}{\textbf{\tiny{\plusvalue{2.9}}}} \\
        SigLIP ViT-SO-14 &  & 384$^2$ & 878.0 & 20.0 & 26.7 & 60.0 & 33.3 & 13.3 & 66.7 & 33.3 & 26.7 & 53.3 & 37.0 \\
        \rowcolor{lightgray} SigLIP ViT-SO-14 & \cmark & 384$^2$ & 878.0 & 26.7 & 33.3 & 53.3 & 26.7 & 13.3 & 80.0 & 40.0 & 26.7 & 46.7 & \textbf{38.5}{\textbf{\tiny{\plusvalue{1.5}}}} \\
        \hline
        
        DFN ViT-H-14 &  & 224$^2$ & 986.1 & 20.0 & 26.7 & 73.3 & 26.7 & 26.7 & 66.7 & 46.7 & 13.3 & 53.3 & 39.3 \\
        \rowcolor{lightgray} DFN ViT-H-14 & \cmark & 224$^2$ & 986.1 & 20.0 & 20.0 & 80.0 & 40.0 & 46.7 & 66.7 & 46.7 & 20.0 & 53.3 & \cellcolor{lightgray}\textbf{43.7}{\textbf{\tiny{\plusvalue{4.4}}}} \\
        DFN ViT-H-14 &  & 378$^2$ & 986.7 & 13.3 & 20.0 & 53.3 & 33.3 & 26.7 & 66.7 & 40.0 & 20.0 & 40.0 & 34.8 \\
        \rowcolor{lightgray} DFN ViT-H-14 & \cmark & 378$^2$ & 986.7 & 26.7 & 26.7 & 53.3 & 33.3 & 26.7 & 73.3 & 26.7 & 13.3 & 60.0 & \textbf{37.8}{\textbf{\tiny{\plusvalue{3.0}}}} \\
    \bottomrule
    \end{tabular}
    \vspace{-10pt}
\end{table}

To validate that our \ours can effectively mitigate the inherent visual capability deficiencies of CLIP models, we first conduct experiment on various existing CLIP models~\citep{radford2021learning, fang2023data, xu2023demystifying, zhai2023sigmoid}.
Despite the variations in image resolution, model size, training data and methodology among these CLIP models, our method consistently enhances their performances on the MMVP-VLM benchmark.
As presented in Table~\ref{tab:clip_sota}, our framework achieves the best performance improvements (\ie, ↑4-7\%) on OpenAI ViT-L-14 and MetaCLIP ViT-H-14, and even on current best-performing DFN ViT-H-14 our framework realizes a performance gain of nearly 3-5\%. 
This fully demonstrates that \ours is both general and effective in enhancing the fine-grained visual perception capabilities of CLIP models.
Notably, through the generative guidance provided by our self-supervised framework that is free from image-text constraints, the perceptual abilities of CLIP models on almost all visual patterns have the potential to be enhanced.

\begin{figure}[t]
  \centering
  \includegraphics[width=1.0\textwidth]{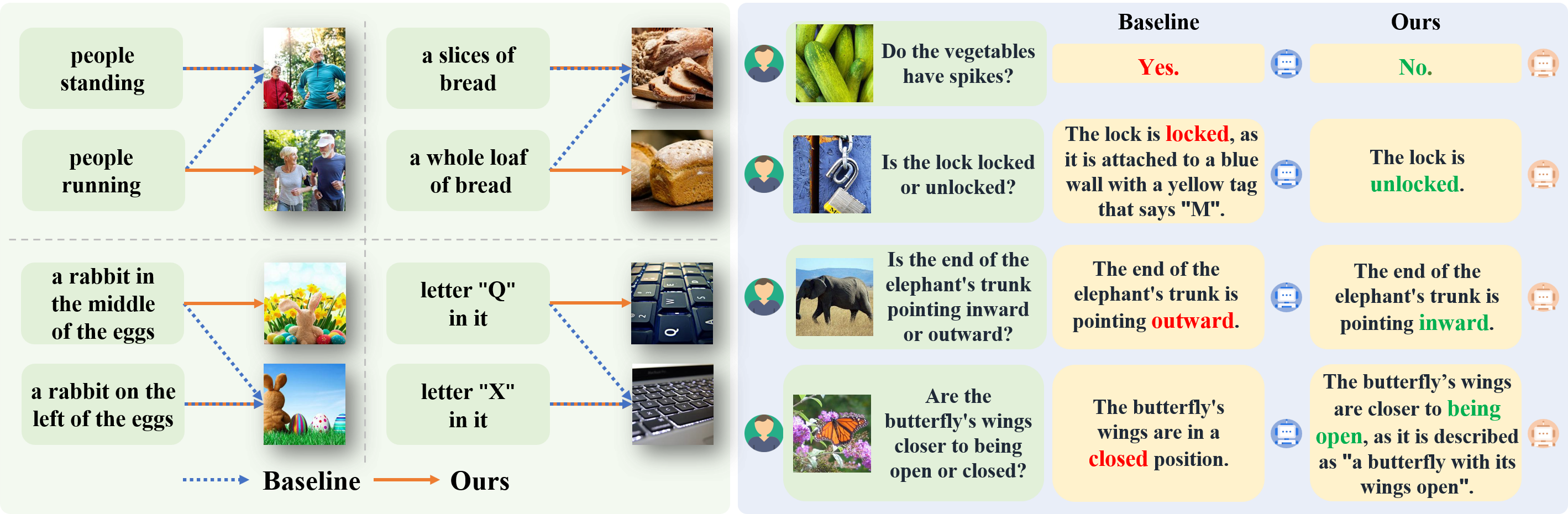}
  \vspace{-15pt}
  \caption{\textbf{Qualitative analysis on MMVP-VLM and MMVP benchmark.}
  \textbf{Left}: The prediction results from the OpenAI ViT-L-14 CLIP before \& after incorporating \ours.
  \textbf{Right}: The prediction results from LLaVA-1.5-7B before \& after using our \ours. The results on both benchmarks show that our framework can greatly enhance CLIP models' fine-grained visual perception capability and effectively alleviate the hallucination problem.} 
  \label{fig:mmvp_vlm}
  \vspace{-15pt}
\end{figure}

\subsection{Backbone Enhancement Performance Evaluation}

Next, with the help of our \ours, we further evaluate the performance gains brought by the enhanced CLIP backbones for multimodal understanding and visual perception tasks.
\vspace{-5pt}

\begin{table}[htbp]
    \caption{\textbf{Performance gains achieved by our enhanced CLIP visual backbone for MLLM (\ie, LLaVA-1.5-7B and LLaVA-1.5-13B) on various V-L understanding tasks.} By refining the CLIP's representation with generative feedback, our method mitigates the visual deficiencies in MLLMs (\ie, LLaVA$^{1.5}$) and improves original instruction following ability.} 
    \label{tab:results_on_mllm_2}
    \centering
    \small 
    \setlength\tabcolsep{1.6pt} 
    \begin{tabular}{l:c:cc:c:ccc:c:cc:c}  
        \toprule
        \multirow{2}{*}{\shortstack{Method}} & \multirow{2}{*}{\shortstack{Ours}} & \multirow{2}{*}{\shortstack{LLM}} & \multirow{2}{*}{\shortstack{Image\\Size}} & \multirow{2}{*}{\shortstack{MMVP}} & \multicolumn{3}{c:}{POPE} & \multirow{2}{*}{\shortstack{MME}} & \multicolumn{2}{c:}{MMBench} 
        & \multirow{2}{*}{\shortstack{LLaVA-Wild}} \\
        & & & & & rand & pop & adv &  & en & cn  & \\
        \hline
        LLaVA$^{1.5}$ &  & Vicuna-7B & 336$^2$ & 24.7 & 87.3 & 86.1 & 84.2 & \textbf{1510.7} & 64.3 & 58.3 & 65.4 \\
        \rowcolor{lightgray} LLaVA$^{1.5}$ & \cmark & Vicuna-7B & 336$^2$ & \textbf{31.3} & \textbf{87.9} & \textbf{87.0} & \textbf{84.6} & 1500.6 & \textbf{66.4} & \textbf{60.6} & \textbf{66.3} \\ 
        \hline
        LLaVA$^{1.5}$ &  & Vicuna-13B & 336$^2$ & 30.7
        & 87.1 & 86.2 & 84.5 & \textbf{1531.3} & 67.7 & \textbf{63.6} & 72.5 \\
        \rowcolor{lightgray} LLaVA$^{1.5}$ & \cmark & Vicuna-13B & 336$^2$ & \textbf{35.3} &\textbf{88.1} & \textbf{87.4} & \textbf{84.8} & 1522.9 & \textbf{69.4} & 63.1 & \textbf{73.5} \\
    \bottomrule
    \end{tabular}
    \vspace{-10pt}
\end{table}

\myparagraph{Enhanced Vision Backbone for MLLMs.}
Firstly, we adopt LLaVA-1.5~\citep{liu2024improved} as the baseline framework to explore the potential of improved visual encoders in MLLM. 
LLaVA employs a pre-trained CLIP vision encoder and trains a projector to semantically align visual tokens with textual tokens from large language model (LLM). 
To ensure fair comparisons, we train our model with the same setting in LLaVA and evaluate model performance on various multimodal understanding benchmarks (\ie, MMVP \citep{tong2024eyes}, 
POPE \citep{li2023evaluating}, MME-Perception \citep{fu2023mme}, MMBench \citep{liu2023mmbench}, MMBench-CN \citep{liu2023mmbench}, LLaVA-Bench-in-the-Wild \citep{liu2024visual}).
It can be clearly seen from Table \ref{tab:results_on_mllm_2} that LLaVA's performance is greatly boosted by replacing the original CLIP vision encoder to ours. 
The big accuracy gains on these benchmarks (except MME) are all thanks to the significant enhancement in CLIP's visual perception capabilities brought by our \ours paradigm utilizing generative feedback.

\begin{table}[htbp]
    \caption{\textbf{Performance gains achieved by our enhanced CLIP backbone with generative guidance on semantic segmentation task.} * denotes the re-implemented results. Boosting CLIP's ability to perceive fine-grained visual details results in considerable benefits for visual dense prediction task.}
    \label{tab:enhanced_vision}
    \centering
    \small
    \setlength\tabcolsep{1.8pt}
    \begin{tabular}{l:c:c:c:c:c:c}
    \toprule
        \multirow{2}{*}{\shortstack{Method}} & \multirow{2}{*}{\shortstack{Backbone}} & \multirow{2}{*}{\shortstack{Ours}}  & \multirow{2}{*}{\shortstack{ADE20K-847}}  & \multirow{2}{*}{\shortstack{ADE20K-150}}  & \multirow{2}{*}{\shortstack{Pascal Context-459}}  & \multirow{2}{*}{\shortstack{Pascal Context-59}} 
         \\
         \\
        \hline
        SAN* & ViT-L-14/224 &  & 10.9 & 29.2 & 14.2 & 55.8 \\
        \rowcolor{lightgray} SAN & ViT-L-14/224 & \cmark & \textbf{11.0} & \textbf{30.2} & \textbf{15.4} & \textbf{56.7} \\
        \hline 
        SAN* & ViT-L-14/336 &  & 11.5 & 30.3 & 14.7 & 56.7 \\
        \rowcolor{lightgray} SAN & ViT-L-14/336 & \cmark & \textbf{11.5} & \textbf{31.8} & \textbf{15.7} & \textbf{57.8} \\
    \bottomrule
    \end{tabular}
\end{table}

\myparagraph{Enhanced Vision Backbone for Fine-Grained Visual Perception.}
We also include segmentation task to evaluate the benefits brought by our enhanced CLIP backbones for visual dense prediction task. 
We adopt the recent state-of-the-art model in the open vocabulary semantic segmentation field, SAN \citep{xu2023side} with CLIP \citep{radford2021learning} at both 224 and 336 image resolutions, as baselines.
Four commonly used benchmarks (\ie, ADE20K-847/150 \citep{zhou2017scene} and Pascal Context-459/59 \citep{mottaghi2014role}
for are employed for performance evaluation.
As shown in Table \ref{tab:enhanced_vision}, with the benefit of our generative tuned CLIP backbone, the baseline models achieve considerable performance improvements on most of segmentation benchmarks and do not suffer performance degradation on the remaining one (\ie, ADE20K-847). 

\vspace{-5pt}
\subsection{Generalization Capability Evaluation}
\label{subsec:generalization}
\vspace{-5pt}

\begin{table}[htbp]
    \caption{\textbf{Summary of zero-shot image classification performance on 27 datasets for evaluating model generalization capability.} O-1 and M-1 separately represent OpenAI ViT-L-14/224 and MetaCLIP ViT-H-14/224. \ours greatly improves CLIP's ability to perceive visual details, while preserving its outstanding generalization capabilities.}
    \label{tab:generalization_results1}
    \centering
    \scriptsize
    \setlength\tabcolsep{0.6pt} 
    \begin{tabular}{c:c:cccccc:ccccccccccccccccccccc:c}
        \toprule
        \rotatebox[origin=l]{90}{{Method}} & \rotatebox[origin=l]{90}{{Ours}} &
        \rotatebox[origin=l]{90}{{ImageNet-1K}} &
        \rotatebox[origin=l]{90}{{ImageNet-V2}} &
        \rotatebox[origin=l]{90}{{ImageNet-Adv.}} &
        \rotatebox[origin=l]{90}{{ImageNet-Ren.}} &
        \rotatebox[origin=l]{90}{{ImageNet-Ske.}} &
        \rotatebox[origin=l]{90}{{ObjectNet}} &
        \rotatebox[origin=l]{90}{{CIFAR-10}} &
        \rotatebox[origin=l]{90}{{CIFAR-100}} & 
        \rotatebox[origin=l]{90}{{MNIST}} & 
        \rotatebox[origin=l]{90}{{Caltech-101}} & 
        \rotatebox[origin=l]{90}{{SUN397}} & 
        \rotatebox[origin=l]{90}{{FGVC Aircraft}} & 
        \rotatebox[origin=l]{90}{{Country-211}} & 
        \rotatebox[origin=l]{90}{{Stanford Cars}} &
        \rotatebox[origin=l]{90}{{Birdsnap}} & 
        \rotatebox[origin=l]{90}{{DTD}} & 
        \rotatebox[origin=l]{90}{{Eurosat}} & 
        \rotatebox[origin=l]{90}{{FER2013}} & 
        \rotatebox[origin=l]{90}{{Flowers-102}} & 
        \rotatebox[origin=l]{90}{{Food-101}} & 
        \rotatebox[origin=l]{90}{{GTSRB}} & 
        \rotatebox[origin=l]{90}{{PCam}} & 
        \rotatebox[origin=l]{90}{{Pets}} & 
        \rotatebox[origin=l]{90}{{Rendered SST2}} & 
        \rotatebox[origin=l]{90}{{Resisc45}} & 
        \rotatebox[origin=l]{90}{{STL10}} & 
        \rotatebox[origin=l]{90}{{VOC2007}} &
        \rotatebox[origin=l]{90}{\textbf{avg. top-1 acc.}}
        \\
        \hline
         O-1 &  & 75.5 &  69.8 & 70.7  & 87.8  & 59.6  & 69.0  &  95.6 & 75.8  & 76.4 & 86.6 &  67.5 &  31.9 & 31.9  &  77.8 & 51.4  &   55.4 &  60.1 & 49.9  &  79.1 &  93.0 &  50.6 &  52.0 & 93.6 & 68.8  & 64.5 & 99.4 & 77.4 & 69.3  \\
         \rowcolor{lightgray}O-1 & \cmark & 75.5  & 69.7  & 70.8  & 87.7  & 59.5  & 69.1  & 95.5  & 76.3  & 76.1  & 86.8  & 67.5 & 31.8  & 31.8 & 77.9 & 51.8 & 55.1 & 60.2 & 49.4 & 78.9  & 93.0 & 50.2 & 53.8 & 93.7  & 67.0 & 64.5 & 99.3 & 77.8 & 69.3 \\        
        \hline
        M-1 &  &  78.5 &  72.1 & 69.6  &  91.8 &  68.1 &  73.6 & 98.3  &  86.7 &  81.3 & 89.1  & 74.1 &  48.2 & 34.7 & 87.2 & 68.5 & 69.8 & 55.6 & 54.9 &  80.7 & 92.5 & 62.3 & 56.1 & 94.2 & 71.0  & 72.6 & 99.4 & 77.6 & 74.4 \\
        \rowcolor{lightgray} M-1 & \cmark &  78.4 &  71.9 &  69.1 & 91.6  & 67.9  &  73.4 &  98.3 &  86.4 &  81.0 & 89.1  &  74.3 & 47.0 &  34.7 & 87.2 & 67.8 & 69.6 & 55.0 & 55.8 & 80.7 & 92.4  &  62.4 & 54.8 & 94.1 & 70.7  & 73.0 & 99.4 & 77.6 & 74.2 \\
        \bottomrule
        \end{tabular}
\end{table}
\vspace{-5pt}

\vspace{-5pt}
\begin{table}[htbp]
    \caption{\textbf{Summary of zero-shot text and image retrieval performance on Flickr30K and COCO datasets for evaluating model generalization capability.} Our \ours significantly enhances CLIP's visual detail perception ability while maintaining its excellent generalization capabilities.}
    \label{tab:generalization_results2}
    \centering
    \small
    \setlength\tabcolsep{0.6pt}
    \begin{tabular}{l:c:c:ccc:ccc:ccc:ccc}
    \toprule
        \multirow{3}{*}{Method} & \multirow{3}{*}{Ours} & \multirow{3}{*}{Image Size} & \multicolumn{6}{c:}{Zero-Shot \textbf{Text} Retrieval} & \multicolumn{6}{c}{Zero-Shot \textbf{Image} Retrieval} \\
        &  &  & \multicolumn{3}{c:}{Flickr30K} & \multicolumn{3}{c:}{COCO} & \multicolumn{3}{c:}{Flickr30K} & \multicolumn{3}{c}{ COCO} \\
    
        &  &  &  R@1 &  R@5 &  R@10 &  R@1 &  R@5 &  R@10 &  R@1 &  R@5 &  R@10 &  R@1 &  R@5 &  R@10 \\
        
        \hline
         OpenAI ViT-L-14 &  &  224$^2$ &  85.1 &  97.3 & 99.0  &  56.4 &  79.4 & 86.6  & 65.2 & 87.3  &  92.0 &  36.5 & 61.0  & 71.1  \\
         \rowcolor{lightgray}OpenAI ViT-L-14 & \cmark  &  224$^2$  & 85.3  &  97.3 &  99.0 &  56.7 & 79.7  & 87.0  & 64.4  & 86.9  &  92.0 &  36.6 &  61.0 &  71.3 \\
         \hline
         MetaCLIP ViT-H-14 &  &  224$^2$ & 89.5  &  98.8 &  99.7 & 65.5  &  85.2 &  91.1 & 76.8 & 93.9  & 96.6  & 48.2  &  72.3 & 81.1  \\
         \rowcolor{lightgray}MetaCLIP ViT-H-14 & \cmark  &  224$^2$  & 89.2  &  98.7 &  99.7 & 65.5  &  85.0 & 91.1 &  77.3 & 93.8  & 96.7 & 48.4 &  72.4 &  81.1 \\
    \bottomrule
    \end{tabular}
    \vspace{-15pt}
\end{table}

After comprehensive validation of our approach's ability to boost CLIP models' fine-grained visual perception abilities, we conduct a thorough assessment of CLIP model's original generalization ability. 
The details about all benchmarks can be found at Table \ref{tab:eval_datasets} in \textcolor{blue}{Appendix}.
Specifically, OpenAI ViT-L-14 \citep{radford2021learning} and MetaCLIP ViT-H-14 \citep{xu2023demystifying}, which are widely used and shows the greatest performance gains on MMVP-VLM benchmark, are adopted as our baselines. 
We present their zero-shot accuracies on 27 image classification benchmarks in Table \ref{tab:generalization_results1}.
It is evident that our \ours significantly enhances CLIP models' representations of fine-grained visual details without adversely affecting the generalization capabilities of the baselines to a large extent. 
Furthermore, Table \ref{tab:generalization_results2} illustrates the comparison of zero-shot image and text retrieval performance before and after incorporating \ours. 
The quantitative results reaffirm that optimizing CLIP models' representations with \ours preserves the original great generalization ability.
Given that these tasks heavily rely on the CLIP visual backbone's global semantic understanding, it is reasonable that our generative guided CLIPs do not achieve much performance improvements on these tasks.
\vspace{-5pt}
\subsection{Ablation Study}
\label{sec:ablation}
\vspace{-5pt}

By taking OpenAI ViT-L-14/224 as the baseline model, we conduct comprehensive ablation studies on MMVP-VLM regarding each pattern and average score, exploring the effect of condition designing, introduced data scale and adopted diffusion models for \ours.

\begin{table}[htbp]
    \caption{\textbf{Ablation study on the condition design for diffusion models.} G and L denote visual class token and patch tokens. Compared to using semantic matching constraints from image-text pairs as condition, taking visual features as condition for representation-level optimization is more effective. Furthermore, an appropriate degree of visual dense recap scheme is also crucial for \ours.} 
    \label{tab:ablation_condition}
    \centering
    \small
    \setlength\tabcolsep{2pt}
    \begin{tabular}{c:c:ccccccccc:c}
    \toprule
         \multirow{2}{*}{\shortstack{Visual Condition}} & \multirow{2}{*}{\shortstack{Textual Condition}} & \multirow{2}{*}{\faCompass} & \multirow{2}{*}{\faSearch} & \multirow{2}{*}{\faSync} & \multirow{2}{*}{\faSortNumericUp} & \multirow{2}{*}{\faMapPin} & \multirow{2}{*}{\faPalette}  & \multirow{2}{*}{\faCogs}  & \multirow{2}{*}{\faFont} & \multirow{2}{*}{\faCamera}  & \multirow{2}{*}{\shortstack{Average}}
         \\
         \\
        \hline 
        \fmark & \fmark & 13.3 & 13.3 & 20.0 & 20.0 & 13.3 & 53.3 & 20.0 & 6.7 & 13.3 & 19.3 \\
        \hline
        \fmark & Real Caption & 6.7 & 13.3 & 26.7 & 20.0 & 6.7 & 53.3 & 33.3 & 13.3 & 20.0 & 21.5{\textbf{\tiny{\plusvalue{2.2}}}} \\
        \fmark & Free-Source Phrase & 6.7 & 13.3 & 20.0 & 20.0 & 6.7 & 53.3 & 26.7 & 13.3 & 26.7 & 20.7{\textbf{\tiny{\plusvalue{1.4}}}} \\
        \hline
        only G  & Empty Caption & 20.0 & 20.0 & 20.0 & 20.0 & 13.3 & 46.7 & 26.7 & 20.0 & 13.3 & 22.2{\textbf{\tiny{\plusvalue{2.9}}}}\\
        \rowcolor{lightgray} G + part\_L  & Empty Caption & 13.3 & 20.0 & 40.0 & 6.7 & 20.0 & 53.3 & 46.7 & 20.0 & 13.3 & \textbf{25.9}{\textbf{\tiny{\plusvalue{6.6}}}}\\
        G + all\_L & Empty Caption & 6.7 & 20.0 & 40.0 & 6.7 & 6.7 & 40.0 & 40.0 & 6.7 & 13.3 & 20.0{\textbf{\tiny{\plusvalue{0.7}}}}\\
    \bottomrule
    \end{tabular}
    \vspace{-6pt}
\end{table}

\myparagraph{Effect of Condition Design for Diffusion Models.} 
We first probe into the influence of diffusion models' condition designing.
As elaborated in Sec. \ref{subsec:diff_condition}, the condition design of diffusion models is vital as it directly determines the upper limit of CLIP models' enhanced representation quality.
As shown in Table \ref{tab:ablation_condition}, we consider two condition settings: 1) using pure text embeddings as condition, which similar to Diffusion-TTA \citep{prabhudesai2023diffusion} (rows 3-4); 2) incorporating densely recapped visual features and empty text's embeddings as condition (rows 5-7).
Specifically, whether using real caption matched with images from CC-3M dataset or using free-source phrases about image details, guiding CLIP through image-text matching constraints can yield performance gains. 
However, since this manner does not originate from the representation level, the achieved gain is not significant. 
In contrast, our condition design introduces appropriately densified visual features, constructing a framework that only uses images to achieve self-supervised optimization of CLIP representations and detaching from image-text form constraints.
\ours helps CLIP achieve the best performance gains (↑6.6\%) by introducing partial visual local features coupled with class tokens as condition. Introducing too few or many local tokens results in visual density being too low (row 5) or high (row 7), both of which reduce the achieved performance improvement.

\vspace{-5pt}
\begin{table}[htbp]
    \caption{\textbf{Ablation study on the data scaling property of our \ours with different data scales.} Training time is measured by \# gpus$\times$hours. \ours demonstrates great potential with data scaling properties, where the increase in data volume proportionally enlarges the performance gains.}
    \label{tab:ablation_datascale}
    \centering
    \small
    \setlength\tabcolsep{2pt}
    \begin{tabular}{c:c:ccccccccc:c}
    \toprule
        \multirow{2}{*}{\shortstack{Data Scale}} & \multirow{2}{*}{\shortstack{Training Time}} & \multirow{2}{*}{\faCompass}  &  \multirow{2}{*}{\faSearch} & \multirow{2}{*}{\faSync} & \multirow{2}{*}{\faSortNumericUp} & \multirow{2}{*}{\faMapPin} & \multirow{2}{*}{\faPalette}  & \multirow{2}{*}{\faCogs}  & \multirow{2}{*}{\faFont} & \multirow{2}{*}{\faCamera}  & \multirow{2}{*}{\shortstack{Average}}
         \\
         \\
        \hline
        \fmark & \fmark & 13.3 & 13.3 & 20.0 & 20.0 & 13.3 & 53.3 & 20.0 & 6.7 & 13.3 & 19.3 \\
        \hline
        25\% & 16.8 & 6.7 & 13.3 & 20.0 & 13.3 & 6.7 & 46.7 & 53.3 & 13.3 & 13.3 & 20.7{\textbf{\tiny{\plusvalue{1.4}}}}\\
        50\% & 32.8 & 13.3 & 13.3 & 40.0 & 13.3 & 6.7 & 26.7 & 53.3 & 20.0 & 13.3 & 22.2{\textbf{\tiny{\plusvalue{2.9}}}}\\
        75\% & 49.6 & 6.7 & 26.7 & 40.0 & 13.3 & 6.7 & 53.3 & 53.3 & 6.7 & 6.7 & 23.7{\textbf{\tiny{\plusvalue{4.4}}}}\\
        \rowcolor{lightgray} 100\% & 66.4 & 13.3 & 20.0 & 40.0 & 6.7 & 20.0 & 53.3 & 46.7 & 20.0 & 13.3 & \textbf{25.9}{\textbf{\tiny{\plusvalue{6.6}}}}\\
    \bottomrule
    \end{tabular}
    \vspace{-8pt}
\end{table}

\myparagraph{Data Scaling Property.} 
Then, we investigate the potential data scaling property of our \ours by taking CC-3M dataset as training data. 
The corresponding results are presented in Table \ref{tab:ablation_datascale}. 
It is obvious that the CLIP model's performance on MMVP-VLM benchmark is consistently improved with more training samples.
As the ratios of introduced samples continue to rise, there's no sign of diminishing gains in accuracy, suggesting that our framework has great potential with continually scaled up training data.
Noticeably, by integrating 100\% training data, our method greatly boosts CLIP's visual perception capability (\ie, approximately ↑7\%) with 66.4 \# gpus $\times$ hours. 
It means if 8 gpus are available, \ours only need 8.3 hours training time to realize considerable performance gains for CLIP models.
Besides, this setting results in an adaptation time of roughly 0.01 seconds per sample, which fully proves the efficacy of our method.

\vspace{-6pt}
\begin{table}[htbp]
    \caption{\textbf{Ablation study on the adopted generative diffusion models in our \ours.} Training time is measured by \# gpus$\times$hours. Our framework is not sensitive to the version of stable diffusion models, consistently brings representation enhancement for CLIP models.}
    \label{tab:ablation_diffusionversion}
    \centering
    \small
    \setlength\tabcolsep{1.2pt}
    \begin{tabular}{c:c:c:ccccccccc:c}
    \toprule
        \multirow{2}{*}{\shortstack{Method}} & \multirow{2}{*}{\shortstack{Diffusion Resolution}} & \multirow{2}{*}{\shortstack{Training Time}} & \multirow{2}{*}{\faCompass}  &  \multirow{2}{*}{\faSearch} & \multirow{2}{*}{\faSync} & \multirow{2}{*}{\faSortNumericUp} & \multirow{2}{*}{\faMapPin} & \multirow{2}{*}{\faPalette}  & \multirow{2}{*}{\faCogs}  & \multirow{2}{*}{\faFont} & \multirow{2}{*}{\faCamera}  & \multirow{2}{*}{\shortstack{Average}}
         \\
         \\
        \hline 
        \fmark & \fmark & \fmark & 13.3 & 13.3 & 20.0 & 20.0 & 13.3 & 53.3 & 20.0 & 6.7 & 13.3 & 19.3 \\
        \hline
        DiT-XL/2 & 512$^2$ & 80.8 & 20.0 & 13.3 & 20.0 & 6.7 & 6.7 & 20.0 & 20.0 & 6.7 & 6.7 & 13.3{\textbf{\tiny{\minusvalue{6.0}}}}\\
        SD-1-4 & 512$^2$ & 71.2 & 20.0 & 13.3 & 26.7 & 20.0 & 13.3 & 40.0 & 33.3 & 26.7 & 13.3 & 23.0{\textbf{\tiny{\plusvalue{3.7}}}}\\
        \rowcolor{lightgray} SD-2-1-base & 512$^2$ & 66.4 & 13.3 & 20.0 & 40.0 & 6.7 & 20.0 & 53.3 & 46.7 & 20.0 & 13.3 & \textbf{25.9}{\textbf{\tiny{\plusvalue{6.6}}}}\\
        SD-xl-base-1.0 & 512$^2$ & 90.4 & 20.0 & 20.0 & 26.7 & 26.7 & 6.7 & 46.7 & 33.3 & 6.7 & 26.7 & 23.7{\textbf{\tiny{\plusvalue{4.4}}}}\\
    \bottomrule
    \end{tabular}
    \vspace{-6pt}
\end{table}

\myparagraph{Effect of Diffusion Model Structures.} 
At last, we explore the effect of incorporating various types of diffusion models for generative guidance.
Specifically, two types of diffusion models are employed as generative guidance in our \ours, including DiT \citep{peebles2023scalable} and stable diffusion (SD) series \citep{rombach2022high} (\ie, 1-4, 2-1-base, xl-base-1.0). 
It is clear in Table \ref{tab:ablation_diffusionversion} that our method achieves the biggest performance gain (\ie, ↑6.6) on MMVP-VLM by integrating SD-2-1-base.
Furthermore, we observe that integrating DiT-XL/2 as generative guidance exacerbates the perceptual ability of original CLIP model in capturing visual details. 
We attribute this to DiT's relatively poor representation quality compared to SD models.
For the included SD series, the quantitative results in Table \ref{tab:ablation_diffusionversion} also demonstrate that \ours is not sensitive to version of SD models, which can consistently refine CLIP's feature representations within our framework.

\vspace{-5pt}
\section{Conclusion}
\vspace{-5pt}

In this work, we focus on addressing the visual limitation of CLIP models that struggle with distinguishing fine-grained image details. 
We present the first work to explore leveraging generative feedback from text-to-image diffusion models to directly optimize CLIP models' representations.
Specifically, by feeding dense visual features from CLIP as condition to the diffusion models and introducing the reconstruction loss from diffusion process onto the CLIP model's optimization, we establish a self-supervised framework \ours.
Notably, this architecture is simple and clean, requiring no additional plugins while demonstrating significant potential. 
Extensive evaluations demonstrate that our \ours not only substantially enhances CLIP models' performance on the MMVP-VLM benchmark that measures visual abilities of vision-language models, but also aids in improving the performance of MLLMs and vision networks respectively on multimodal and visual understanding tasks.
Furthermore, experiments on 29 benchmarks evaluating generalization capabilities confirm that our self-supervised \ours maintains the CLIP models' original excellent generalization capabilities.

\vspace{-5pt}
\section{Limitations and Future Topics}
\vspace{-5pt}

One potential limitation of this work is that the data scale for generative fine-tuning and the model capacity of our \ours could be scaled up further to push better CLIP representations and performance. 
Moreover, this work mainly focuses on designing a simple but effective framework to enhance CLIP models with generative diffusion process, which means although our \ours demonstrates the newly exploited potential of using generative diffusion models for better CLIP models' representation guidance, it can be integrated with finer-grained supervision scheme to further boost discriminative model capabilities. 
Furthermore, exploring additional modalities beyond image-text data, \eg, video and audio, is also a promising direction for investigation.
Since this work is just a beginning in this direction, it opens up a future research perspective to develop a more general and powerful framework based on diffusion models that could enhance vision-language foundation models.

\clearpage

{\small
\bibliographystyle{iclr2024_conference}
\bibliography{iclr2024_conference}
}

\newpage
\appendix
\section*{Appendix}

\section{Datasets for Generalization Capability Evaluation}

\begin{table}[htbp]
    \caption{\textbf{Datasets used to evaluate CLIP models's generalization capability}.} 
    \label{tab:eval_datasets}
    \centering
    \small 
    \setlength\tabcolsep{2pt}
    \begin{tabular}{lllr}  
    \toprule
     Dataset & Classes & Test size & Evaluation Metric \\
    \midrule
     ImageNet-1K~\citep{deng2009imagenet} &  1000 &  50,000 &  accuracy \\
     ImageNet-V2~\citep{recht2019imagenetv2} &  1000 &  10,000 &  accuracy \\
     ImageNet-Adversarial~\citep{inadv} &  1000 &  7,500 &  accuracy \\
     ImageNet-R(endition)~\citep{inren} &  1000 &  30,000 &  accuracy \\
     ImageNet-Sketch~\citep{inske} &  1000 &  50,899 &  accuracy \\
     ObjectNet~\citep{objectnet} &  1000 &  50,273 &  accuracy \\
     CIFAR-10~\citep{krizhevsky2009learning} &  10 &  10,000 &  accuracy \\
     CIFAR-100~\citep{krizhevsky2009learning} &  100 &  10,000 &  accuracy \\
     MNIST~\citep{lecun1998gradient} &  10 &  10,000 &  accuracy \\
     Caltech101~\citep{fei2004learning} &  101 &  9144 &  accuracy \\
     SUN397~\citep{xiao2010sun} &  397 &  108,754 &  accuracy \\
     FGVC Aircraft~\citep{maji2013fine} &  100 &  3,333 &  accuracy \\
     Country-211~\citep{clip} &  211 &  21,100 &  accuracy \\
     Stanford Cars~\citep{krause20133d} &  196 &  8,041 &  accuracy \\
     Birdsnap~\citep{berg2014birdsnap} &  500 &  2,195 &  accuracy \\
     Describable  Textures~\citep{cimpoi14describing} &  47 &  1,880 &  accuracy \\
     EuroSAT\citep{helber2019eurosat} &  10 &  27,000 &  accuracy \\
     Facial Emotion Recognition 2013~\citep{goodfellow2013challenges} &  8 &  3,574 &  accuracy \\
     Oxford Flowers 102~\citep{nilsback2008automated} &  102 &  6,149 &  accuracy \\
     Food-101~\citep{bossard2014food} &  102 &  25,250 &  accuracy \\
     GTSRB~\citep{stallkamp2012man} &  43 &  12,630 &  accuracy \\
     PatchCamelyon~\citep{veeling2018rotation} &  2 &  32,768 &  accuracy \\
     Oxford-IIIT Pets~\citep{parkhi12a} &  37 &  3,669 &  accuracy \\
     Rendered SST2~\citep{clip} &  2 &  1,821 &  accuracy \\
     RESISC45~\citep{cheng2017remote} &  45 &  31,500 &  accuracy \\
     STL-10~\citep{coates2011analysis} &  10 &  8000 &  accuracy \\
     Pascal VOC 2007 Classification~\citep{pascal-voc-2007} &  20 &  4,952 &  accuracy \\
     Flickr30K~\citep{flickr30K} &  - &  1000 &  recall \\
     COCO~\citep{lin2014coco} &  - &  5000 &  recall \\
    \bottomrule
    \end{tabular}
\end{table}

\section{Pseudo code for \ours Pipeline}

\begin{algorithm}[htbp]
   \caption{$\texttt{\ours}$}
   \label{alg:diff-tta}
\begin{algorithmic}[1]
    \STATE {\bfseries Input:} Image $x$, CLIP model weights $\theta$, 
    diffusion model weights $\phi$, representation optimization steps $N$, batch size $B$, learning rate $\eta$, optimized CLIP model weights $\theta^*$. 
    \FOR{optimization step $s \in (1, \dots, N)$}
        \STATE Compute current CLIP's visual features $\theta(x)$ as partial condition $\cond$
        \STATE Sample timesteps $\{t_i\}_{i=1}^B$ and noises $\{\epsilon_i\}_{i=1}^B$
        \STATE Loss $L(\theta, \phi) = \frac{1}{N} \sum_{i = 1}^B \| \epsilon_{\phi}(\sqrt{\bar{\alpha}_{t_i}} x + \sqrt{1-\bar{\alpha}_{t_i}} \epsilon_i, \cond , t_i) - \epsilon_i \|^2$
        \STATE Update CLIP weights $\theta^* \leftarrow \theta - \eta \nabla_\theta L(\theta, \phi)$
    \ENDFOR
    \RETURN optimized CLIP weights $\theta^*$
\end{algorithmic}
\end{algorithm}

\clearpage

\end{document}

%% file: math_commands.tex
\usepackage{amsmath,amsfonts,bm}

\def\eqref#1{equation~\ref{#1}}

\def\1{\bm{1}}

\DeclareMathAlphabet{\mathsfit}{\encodingdefault}{\sfdefault}{m}{sl}
\SetMathAlphabet{\mathsfit}{bold}{\encodingdefault}{\sfdefault}{bx}{n}